\documentclass[11pt]{article}

\PassOptionsToPackage{table}{xcolor}
\usepackage[final]{acl}

\usepackage{times}
\usepackage{latexsym}
\usepackage[T1]{fontenc}
\usepackage[utf8]{inputenc}
\usepackage{microtype}
\usepackage{inconsolata}

\usepackage{graphicx}
\usepackage{booktabs}
\usepackage{enumitem}
\usepackage[most]{tcolorbox}
\usepackage{amsmath}
\usepackage{xfp}
\usepackage{multirow}

\definecolor{topicblue}{RGB}{52, 144, 220}
\definecolor{lightgray}{gray}{0.9}
\definecolor{darkgreen}{rgb}{0.0, 0.5, 0.0}
\definecolor{darkred}{rgb}{0.8, 0.0, 0.0}

\newcommand{\gainavg}[2]{%
  #2%
  \ifdim \fpeval{#2 > #1}pt > 0pt
    ~{\textcolor{darkgreen}{\scriptsize(+\fpeval{round(#2-#1,2)})}}%
  \else
    \ifdim \fpeval{#2 < #1}pt > 0pt
      ~{\textcolor{darkred}{\scriptsize(\fpeval{round(#2-#1,2)})}}%
    \fi
  \fi
}
\newcommand{\bgainavg}[2]{%
  \textbf{#2}%
  \ifdim \fpeval{#2 > #1}pt > 0pt
    ~{\textcolor{darkgreen}{\scriptsize(+\fpeval{round(#2-#1,2)})}}%
  \else
    \ifdim \fpeval{#2 < #1}pt > 0pt
      ~{\textcolor{darkred}{\scriptsize(\fpeval{round(#2-#1,2)})}}%
    \fi
  \fi
}

\begin{document}

\title{Does Understanding Inform Generation in Unified Multimodal Models? \\ From Analysis to Path Forward}

\author{
Yuwei Niu$^{1,3,*}$, Weiyang Jin$^{4,*}$, Jiaqi Liao$^{5}$, Chaoran Feng$^{1}$, Peng Jin$^{6}$, \\
\bfseries Bin Lin$^{1}$, Zongjian Li$^{1}$, Bin Zhu$^{1}$, Weihao Yu$^{1,2}$, Li Yuan$^{1,2,\dagger}$ \\
$^{1}$Shenzhen Graduate School, Peking University \quad
$^{2}$PengCheng Laboratory \quad
$^{3}$Chongqing University \\
$^{4}$The University of Hong Kong \quad
$^{5}$The Hong Kong University of Science and Technology \quad
$^{6}$Qwen Team \\
$^{*}$Equal contribution. \quad $^{\dagger}$Corresponding author. \\
\texttt{niuyuwei04@gmail.com, yuanli-ece@pku.edu.cn}
}

\maketitle

\begin{abstract}
  Recent years have witnessed significant progress in Unified Multimodal Models, yet a fundamental question remains: Does understanding truly inform generation in Unified Multimodal Models? To investigate this, we introduce \textbf{UniSandbox}, a decoupled evaluation framework paired with controlled, synthetic datasets to avoid data leakage and enable detailed analysis. Our findings reveal a significant understanding-generation gap, which is mainly reflected in two key dimensions: reasoning generation and knowledge transfer. Specifically, for reasoning generation tasks, we observe that explicit Chain-of-Thought (CoT) in the understanding module effectively bridges the gap, and further demonstrate that a self-training approach can successfully internalize this ability, enabling implicit reasoning during generation. Additionally, for knowledge transfer tasks, we find that CoT assists the generative process by helping retrieve newly learned knowledge, and also discover that query-based architectures inherently exhibit latent CoT-like properties that affect this transfer. UniSandbox provides preliminary insights for designing future unified architectures and training strategies that truly bridge the understanding-generation gap.
\end{abstract}

\section{Introduction}

Recent years have witnessed significant progress in Unified Multimodal Understanding and Generation Models~\citep{vilau2024, janus2024, dongdreamllm,liao2025langbridge, fang2024puma, show-o,lin2025uniworld,li2025uniworld,li2025manzano,wu2025openuni}, with powerful architectures designed to integrate both tasks. Yet this trend raises a more fundamental question: what is the true synergy gained by housing both capabilities in a single model? We investigate one crucial direction: whether a model's rich language priors can genuinely inform and guide its visual generation.

Existing works~\citep{tong2024metamorph,sun2025t2i,xie2025mme,zou2025uni,li2025gir}, such as WISE~\citep{niu2025wise}, aim to evaluate whether models can use world knowledge for visual generation in complex contexts. For instance, given the prompt ``A flower commonly gifted on Mother's Day,'' an ideal model would first deduce ``carnation'' from its world knowledge, and this concept would then serve as a clear instruction for the visual generator, demonstrating a true synergy in which understanding guides generation.

However, while these works valuably reveal that models perform poorly when required to use internal knowledge or reasoning for generation, they conflate multiple failure modes and offer no fine-grained attribution. We cannot tell whether a failure stems from a \textit{knowledge} deficit (the model lacks the facts), a \textit{reasoning} deficit (it cannot deduce the answer from known facts and rules), or a \textit{transfer} failure (the understanding module knows the answer, but the generation module cannot use it). Moreover, potential data leakage renders the results unreliable, as a model's ``success'' might stem from memorizing training pairs rather than genuine reasoning. This dual problem of non-attributability and unreliability makes it hard to identify which architectures or training strategies best foster understanding-generation synergy.

To analyze the understanding-generation gap and address the above issues, we introduce \textbf{UniSandbox}\footnote{Code and data are available at \url{https://github.com/PKU-YuanGroup/UniSandBox}.}, a decoupled framework with synthetic, low-leakage data. UniSandbox isolates understanding into two dimensions, \textit{Knowledge} and \textit{Reasoning}, allowing us to precisely attribute model performance beyond data memorization, and we use it to evaluate several state-of-the-art unified models.

For reasoning generation, we design reasoning-driven tasks that require generating images from mathematical calculation or logical deduction. Without explicit Chain-of-Thought (CoT), all open-source models score near zero, indicating that their generation degenerates into a shallow word-to-pixel mapping, essentially a ``bag-of-words''~\citep{yuksekgonul2022and}. With CoT, the understanding module solves the problem and explicitly guides generation. We further explore whether a simple self-training method can internalize this capability without explicit CoT, turning explicit reasoning implicit and offering a direction for optimizing training paradigms.

In parallel, for knowledge transfer, we inject new knowledge into the understanding module and test whether the generation module can reproduce it through image generation. Mainstream architectures fail this task, while CoT enables internal knowledge retrieval that transfers the knowledge to the generation module, substantially improving performance. Visualization analysis further reveals that query-based architectures, which use extra queries to extract information as the generation condition, exhibit a latent CoT-like capability.

The contributions of our work are as follows:
\begin{itemize}
    \item \textbf{The UniSandbox Framework.} We propose UniSandbox, a novel decoupled evaluation framework for investigating whether understanding truly informs generation. Through synthetic data and fine-grained analysis, our framework reveals a significant understanding-generation gap in existing models.
    \item \textbf{Revealing Flaws in Reasoning Generation.} We confirm the severe deficiency of existing models in reasoning-based generation, demonstrate the role of Chain-of-Thought as an explicit ``bridge", and explore how this capability can be internalized through self-training.
    \item \textbf{Investigating the Knowledge Transfer Bottleneck.} We confirm that models struggle to transfer newly acquired knowledge to the generation module, and again identify CoT as an effective activator. Furthermore, we find that query-based architectures exhibit a ``latent knowledge transfer capability", providing a key design reference for future UMMs.
\end{itemize}

\section{UniSandbox: Controlled Evaluation}
To investigate whether and how the ``understanding'' capabilities of Unified Multimodal Models benefit their ``generation'' tasks, we decompose ``understanding'' into two core dimensions: \textit{Knowledge} (the memory and retrieval of factual information) and \textit{Reasoning} (the application of procedural rules). This enables fine-grained attribution, pinpointing whether a model's failure stems from an inability to invoke newly acquired knowledge or to apply logical reasoning when guiding image generation. To reduce the risk of data contamination from overlap between pre-training and evaluation data, we build our training and evaluation pipelines on controlled synthetic data designed to minimize such overlap, providing a ``sandbox environment'' for analysis.

We selected a representative set of models spanning three paradigms: (1) Autoregressive (AR), e.g., Janus-Pro-7B~\citep{januspro2025}; (2) AR + Diffusion (shallow fusion), including Qwen-Image~\citep{wu2025qwen}, which uses the last hidden state as the generation condition, and Blip3o~\citep{chen2025blip3}, which uses queries to extract conditional information; and (3) AR + Diffusion (deep fusion), e.g., BAGEL~\citep{deng2025emerging}, which deeply integrates understanding and generation in one transformer.
\section{Reasoning Generation}
We first focus on the ``reasoning'' dimension to explore whether the model's internal logic can guide generation, designing two families of synthetic tasks: \textbf{Mathematical Operations} and \textbf{Symbolic Mapping}. These serve as ideal probes for core reasoning: both readily produce out-of-distribution data, their difficulty scales precisely with the number of computational steps or the mapping depth, and they compel the model to move beyond superficial semantics to abstract, procedural symbol manipulation. This lets us quantify the extent to which the generative module inherits the reasoning capabilities of its understanding module. All task instances were generated using GPT-4o, with detailed prompts in Appendix~\ref{app:data}.

\begin{figure*}[t!]
    \centering
    \includegraphics[width=\linewidth]{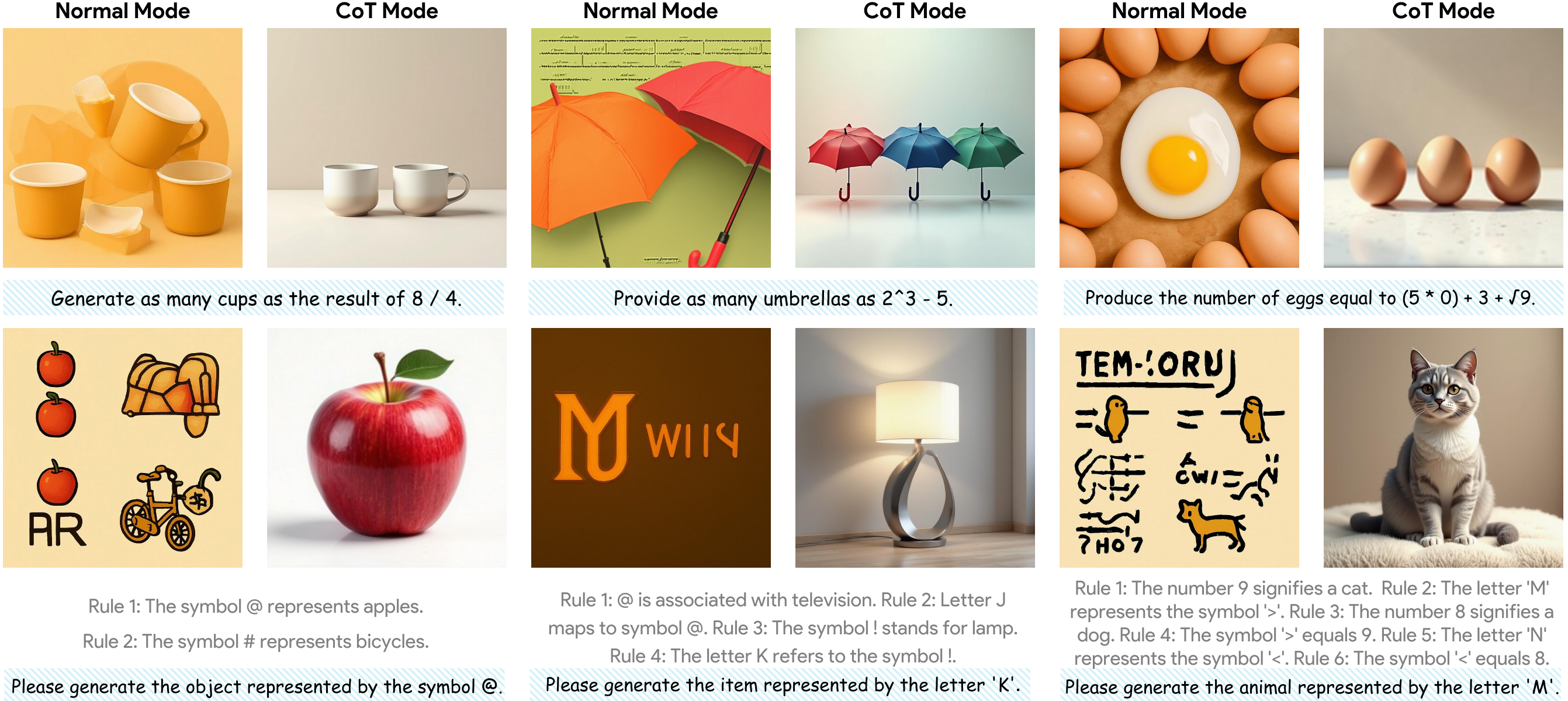}
    
    \caption{Data examples for reasoning generation. All images are generated by BAGEL. Normal and CoT represent generation without and with think mode (Chain-of-Thought mode), respectively. We also show the corresponding prompts.}
    \label{fig:exg}
\end{figure*}

\subsection{Task Design: Explore Reasoning Capabilities}
\subsubsection{Mathematical Operations}

Unlike previous benchmarks~\citep{ghosh2023geneval,huang2023t2i} that test a model's basic generative capacity by directly specifying the number of objects, we require the model to perform sequential calculations and then generate a corresponding number of objects. For instance, with a prompt like ``Provide the same number of erasers as calculated by 3 - 2," the model must first understand and execute the 3 - 2 operation to derive the answer 1, and then use that internally-reasoned result to guide the visual generation of one eraser. This design compels the model to engage its internal reasoning and comprehension abilities before executing the generation task. The difficulty of our tasks progressively increases by extending the length of the operation chain. We use a variety of basic middle-school-level operations, including addition, subtraction, multiplication, division, exponentiation, and modulo operations. All final results are constrained to be integers less than seven, ensuring that the model's potential inability to generate a large number of objects does not interfere with the evaluation. The tasks are structured across three levels of difficulty: first-level tasks require a single arithmetic calculation, while higher-level tasks demand sequential deductions involving two or three distinct, composite operations. Each of the three levels contains 100 prompts, for a total of 300; see \autoref{fig:exg} for examples.

\subsubsection{Symbolic Mapping}

The symbolic mapping tasks are designed to evaluate a model's ability to follow novel, arbitrary rules by reasoning through mapping chains of varying lengths. Unlike a simple text-to-image task that requires direct mapping (e.g., ``apple" to the image of the apple), our method compels the model to reason through a set of given mapping rules, requiring it to identify and generate a unique target object from two given options. For instance, a task might be defined by a set of arbitrary rules: ``The letter `A' represents the number 1, and the number 1 represents a cat." Instead of directly asking for a cat, we prompt the model to generate the animal represented by the letter `A.' To succeed, the model must autonomously reason through a two-step chain—from `A' to `1,' and then from `1' to `cat'—before it can generate the correct image. This design compels the model to leverage its internal reasoning to bridge the gap between abstract symbols and a concrete visual output. We progressively increase the difficulty by extending the length of the mapping chain, from a one-step mapping in first-level tasks to two or three-step chains in higher-level ones. To prevent models from succeeding by random guessing, we employ paired prompts, where each pair is based on the same set of rules but asks for a different result. We generated 100 prompt pairs for each of the three difficulty levels, for a total of 600 individual prompts; see \autoref{fig:exg} for examples.

\subsection{Evaluation Protocol}
\label{sec:eval_protocol}
To avoid the judge biases of using Multimodal Large Language Models (MLLMs) as direct visual judges~\citep{chen2025multi}, we use a two-stage protocol: an MLLM first captions each generated image, and then compares the caption with the ground-truth answer without access to the image, awarding a score of 1 if consistent and 0 otherwise. For symbolic mapping, we apply a stricter criterion: a trial is successful only if the model correctly generates the output for \textbf{both} prompts in a given pair, ensuring that success is attributable to genuine reasoning rather than random guessing. Evaluation details are in Appendix~\ref{app:eval}.

\begin{figure*}[h]
    \centering
    \includegraphics[width=1.0\textwidth]{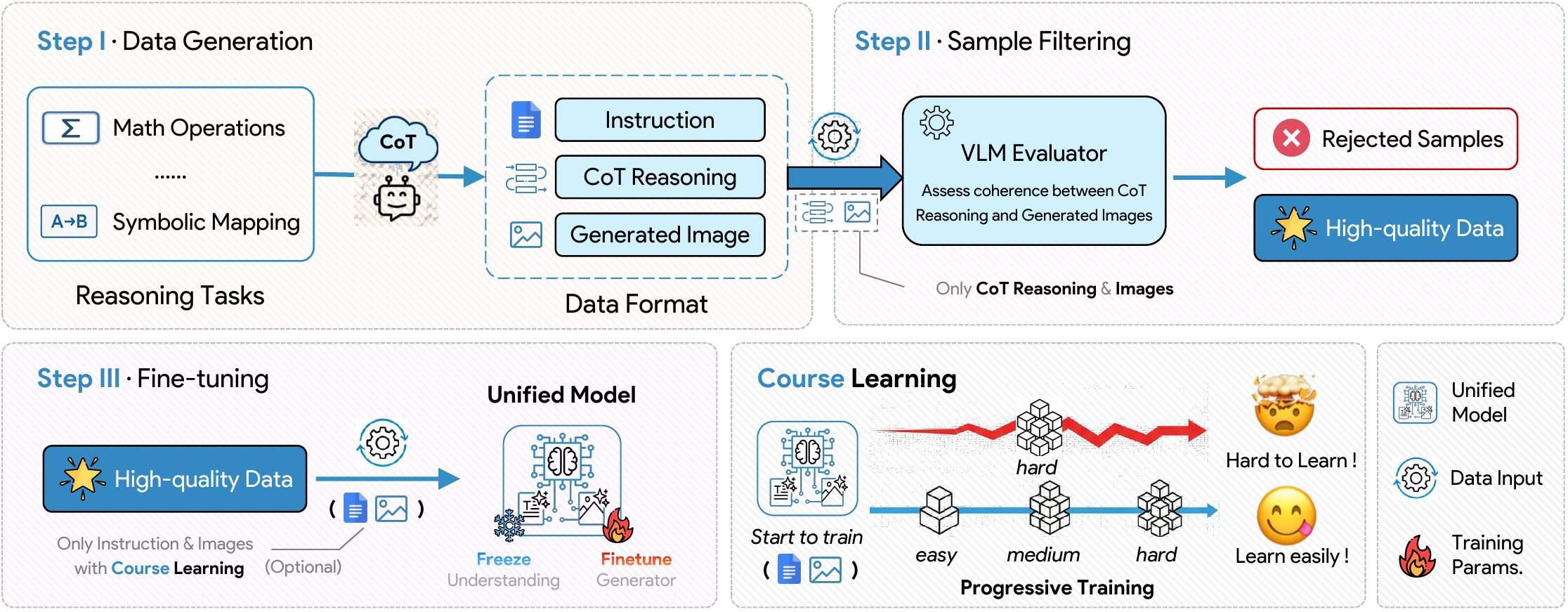}
    \caption{\textbf{Overview of the STARS framework.} 
    It illustrates the three sequential stages: (I) Data Generation, where CoT is leveraged to create reasoning-generation pairs; (II) Sample Filtering, which uses the understanding module of UMMs to curate high-quality data; and (III) Fine-tuning, where the unified model is trained with the filtered data to distill CoT reasoning into its standard generation process.}
    \label{fig:overview}
\end{figure*}

\begin{table*}[htbp]
    \centering
    \caption{Comparison of model performance on the Math and Mapping tasks. Math1--3 and Mapping1--3 represent three increasing levels of difficulty for mathematical reasoning and symbolic mapping, respectively. \textbf{Bold values} indicate the best performance in each column.}
    \label{tab:model_performance}
  \resizebox{0.9\textwidth}{!}{%
    \begin{tabular}{lccccccc}
        \toprule
        \textbf{Model} & \textbf{Math1} & \textbf{Math2} & \textbf{Math3} & \textbf{Mapping1} & \textbf{Mapping2} & \textbf{Mapping3} & \textbf{Average} \\
        \midrule
        Janus-Pro-7B & 0.04 & 0.03 & 0.03 & 0.00 & 0.00 & 0.00 & 0.0167 \\
        Blip3o &0.03 & 0.02 & 0.02 &0.03&0.00&0.00 & 0.0167\\
        Qwen-Image & 0.13 & 0.07 & 0.10 & 0.00 & 0.00 & 0.00 & 0.0500 \\
        BAGEL & 0.07 & 0.06 & 0.04 & 0.00 & 0.00 & 0.00 & 0.0283 \\
        BAGEL + CoT & \textbf{0.60} & \textbf{0.44} & \textbf{0.23} & \textbf{0.75} & \textbf{0.57} & \textbf{0.46} & \textbf{0.5083} \\
        \bottomrule
    \end{tabular}
    }

\end{table*}

\subsection{From Reasoning Failure to CoT Activation}
The results, presented in Table~\ref{tab:model_performance}, reveal a pervasive deficiency. Most models, particularly the open-source ones, show severe limitations on the mathematical operation and symbolic mapping tasks that require logical reasoning: their performance is consistently near zero, and on the more abstract symbolic mapping tasks they fail entirely. This supports our central thesis that the generative component of current unified models operates as a shallow keyword-mapping system. While capable of mapping simple keywords (e.g., generating an apple from the word ``apple''), their generation breaks down when internal, procedural logic is required (e.g., first computing ``3+2" and then generating ``5 apples"). This indicates that the model's internal comprehension is not effectively translated into generative action when logical reasoning is demanded.

However, a key comparative result points to a path toward bridging this gap. When Chain-of-Thought (CoT) is applied to BAGEL, its average score rises substantially, from 0.0283 (BAGEL) to 0.5083 (BAGEL + CoT). This provides strong evidence that \textbf{CoT serves as a crucial mechanism for activating the model's latent language priors and guiding them into the visual generation process.} CoT compels the model to first articulate the reasoning steps and derive a conclusion in the language space, then use this explicit outcome as the generative instruction, effectively transforming a complex logical problem into a straightforward text-to-image task.

\subsection{Can Reasoning Capabilities Be Internalized?}
The above experiments demonstrate that CoT is a key mechanism for activating a model's reasoning capabilities and applying them to generative tasks. This finding naturally raises a central question: can the explicit reasoning processes exhibited in CoT mode be \textbf{internalized} into the model's standard, non-CoT generation process through self-training, thereby enabling it to fundamentally overcome its reasoning generation limitations?

Our framework builds on two key observations.

\begin{itemize}

    \item \textbf{CoT as a ``Teacher":} As shown in the preceding section, CoT effectively guides the unified model ($U_{CoT}$) to perform correct logical reasoning, yielding high-quality reasoning-generation pairs that serve as a reliable source of ``teacher'' signals for fine-tuning.

    \item \textbf{Inherent Self-Verification Capability:} The unified model's powerful visual understanding endows it with an inherent capability to evaluate its own generative output. This allows the model to accurately assess the semantic consistency between a generated image and a textual instruction. In other words, the understanding module ($U_{Ver}$) can serve as its own ``verifier" to determine if its output faithfully adheres to the prompt, as has been demonstrated by prior works~\citep{xie2025reconstructionalignmentimprovesunified,jin2025srum,mao2025unirl}.
\end{itemize}

Motivated by these insights, we propose a preliminary framework named \textbf{\textcolor{topicblue}S}elf-\textbf{\textcolor{topicblue}T}r\textbf{\textcolor{topicblue}a}ining with \textbf{\textcolor{topicblue}R}ejection \textbf{\textcolor{topicblue}S}ampling (\textbf{STARS}). Rather than a mature solution, it is part of the UniSandbox reasoning evaluation, exploring whether models can internalize explicit reasoning to aid their implicit reasoning during generation.

\begin{enumerate}

    \item \textbf{Generation:} First, we leverage the model in CoT mode to generate training samples, $D_{gen}$, for a set of input instructions $I$ from various reasoning tasks (e.g., mathematical operations, symbolic mapping). Each sample in this dataset contains the instruction, the corresponding CoT reasoning trace $C$, and the final output image $O$.
    \begin{equation}
            D_{gen} = \{(I_i, C_i, O_i) | (C_i, O_i) = U_{CoT}(I_i)\}
    \label{eq:data_gen}
\end{equation}
    Each tuple $(I_i, C_i, O_i)$ thus represents a complete instruction-thought-output chain.
    \item \textbf{Filtering:} Next, we employ the model's own understanding faculty, $U_{Ver}$, as a discriminator to filter $D_{gen}$. We define a \textbf{binary consistency score} $S(I, O) = U_{Ver}(I, O) \in \{0, 1\}$, where 1 signifies that the output $O$ is consistent with the instruction $I$. Only samples with a score of 1 are retained. This process yields a curated, high-quality dataset, $D_{fil}$.

\begin{equation}
\begin{aligned}
D_{fil} = \{(I_i, O_i) \mid\;& (I_i, C_i, O_i) \in D_{gen} \\
& {} \land S(I_i, O_i) = 1 \}
\end{aligned}
\label{eq:filtering}
\end{equation}

    Notably, the intermediate reasoning trace $C$ is intentionally discarded at this stage, keeping only the $(I, O)$ pairs.

    \item \textbf{Fine-tuning:}   Finally, the filtered dataset $D_{fil}$ is used to fine-tune the original model's standard generator, $U_{Gen}$, whose parameters are denoted by $\theta$. The objective is to train the model to map instructions directly to the verified outputs, thereby implicitly encoding the logical steps of CoT into the model's parameters.
    \begin{equation}
    \theta' \leftarrow \arg \min_{\theta} \sum_{(I, O) \in D_{fil}}\mathcal{L}(U_{Gen}(I; \theta), O)
    \label{eq:loss}
\end{equation}
    Here, $\mathcal{L}$ is the loss function, and $\theta'$ represents the updated model parameters.
\end{enumerate}

Through this framework, we aim to distill the reasoning process of CoT into the model's standard generative behavior, so that after self-training it can intrinsically reason to guide its visual output even without explicit Chain-of-Thought.

\begin{table*}[t!]
\centering
\small
\caption{Performance of BAGEL trained with STARS on mathematical operations of varying difficulties. Math1--3 represent three increasing levels of difficulty for mathematical operations. Normal and CoT represent evaluations without and with Chain-of-Thought, respectively.}
\label{tab:math}
\begin{tabular}{lccccccc}
\toprule
\textbf{Model} & \multicolumn{3}{c}{\textbf{Normal}} & \multicolumn{3}{c}{\textbf{CoT}} & \textbf{Average} \\
\cmidrule(lr){2-4} \cmidrule(lr){5-7}
& \textbf{Math1} & \textbf{Math2} & \textbf{Math3} & \textbf{Math1} & \textbf{Math2} & \textbf{Math3} & \\
\midrule
BAGEL & 0.07 & 0.06 & 0.04 & 0.60 & 0.44 & 0.23 & 0.24 \\
\midrule
STARS on Math1 & 0.29 & 0.27 & 0.16 & 0.60 & 0.42 & 0.27 & \gainavg{0.24}{0.34} \\
STARS on Math2 & 0.17 & 0.26 & 0.12 & 0.57 & 0.41 & 0.25 & \gainavg{0.24}{0.30} \\
STARS on Math3 & 0.26 & 0.26 & 0.16 & 0.55 & 0.43 & 0.29 & \gainavg{0.24}{0.33} \\
\bottomrule
\end{tabular}
\end{table*}

\subsubsection{STARS on Mathematical Operations}

We first applied the STARS framework to the mathematical operations tasks. For each of the three difficulty levels, we constructed a training set by curating 5,000 high-quality samples generated via CoT and filtered with rejection sampling. The model was then fine-tuned on each of these datasets for 6 epochs with a learning rate of $2 \times 10^{-5}$.

The results in Table~\ref{tab:math} demonstrate the effectiveness of our self-training framework: the model's standard (non-CoT) performance, previously near zero, improves substantially regardless of the training data's difficulty. Moreover, the results reveal strong \textbf{bidirectional cross-difficulty generalization}---training on first-level data improves the harder second- and third-level tasks, and vice versa. This suggests that STARS does not merely memorize task-specific solutions but distills the abstract reasoning principles of CoT into a more robust, generalizable capability. We also provide an ablation on rejection sampling in Appendix~\ref{app:abl}.

\subsubsection{STARS on Symbolic Mapping}

\begin{table*}[t]
\centering
\small
\caption{Performance of BAGEL trained with STARS on symbolic mapping datasets of varying difficulties. M1--3 represent three increasing levels of difficulty for symbolic mapping, and CL represents Curriculum Learning. Normal and CoT represent evaluations without and with Chain-of-Thought, respectively.}
\label{tab:mapping}
\begin{tabular}{@{}llccccccc@{}}
\toprule
\multicolumn{2}{c}{\multirow{2}{*}{\textbf{Model}}} & \multicolumn{3}{c}{\textbf{Normal}} & \multicolumn{3}{c}{\textbf{CoT}} & \multirow{2}{*}{\textbf{Average}} \\ 
\cmidrule(lr){3-5} \cmidrule(lr){6-8}
\multicolumn{2}{c}{} & \textbf{M1} & \textbf{M2} & \textbf{M3} & \textbf{M1} & \textbf{M2} & \textbf{M3} & \\ 
\midrule
\multicolumn{2}{l}{\cellcolor{lightgray}BAGEL} & \cellcolor{lightgray}0 & \cellcolor{lightgray}0 & \cellcolor{lightgray}0 & \cellcolor{lightgray}0.75 & \cellcolor{lightgray}0.57 & \cellcolor{lightgray}0.46 & \cellcolor{lightgray}0.30\\ 
\midrule

\multirow{1}{*}{STARS on M1} &  & 0.69 & 0.10 & 0.10 & 0.66 & 0.39 & 0.38 & \bgainavg{0.30}{0.39} \\
\midrule
\multirow{3}{*}{STARS on M2} & Round-1 & 0.04 & 0.05 & 0.04 & 0.70 & 0.18 & 0.13 & \gainavg{0.30}{0.19} \\
& Round-2 & 0.02 & 0.09 & 0.05 & 0.39 & 0.18 & 0.09 & \gainavg{0.30}{0.14} \\
& Round-3 & 0.04 & 0.00 & 0.00 & 0.63 & 0.00 & 0.00 & \gainavg{0.30}{0.11} \\
\midrule
\multirow{3}{*}{STARS on M3} & Round-1 & 0.11 & 0.06 & 0.01 & 0.72 & 0.23 & 0.28 & \gainavg{0.30}{0.24} \\
& Round-2 & 0.01 & 0.05 & 0.02 & 0.63 & 0.25 & 0.17 & \gainavg{0.30}{0.19} \\
& Round-3 & 0.02 & 0.05 & 0.02 & 0.59 & 0.23 & 0.16 & \gainavg{0.30}{0.18} \\
\midrule
\multirow{3}{*}{STARS with CL} & Round-1 & 0.69 & 0.10 & 0.10 & 0.66 & 0.39 & 0.38 & \gainavg{0.30}{0.39} \\
& Round-2 & 0.61 & 0.47 & 0.22 & 0.68 & 0.62 & 0.40 & \gainavg{0.30}{0.50} \\
& Round-3 & 0.64 & 0.46 & 0.27 & 0.75 & 0.65 & 0.50 & \gainavg{0.30}{0.55} \\
\bottomrule
\end{tabular}
\end{table*}

\begin{figure*}[t]
    \centering
\includegraphics[width=1.0\textwidth]{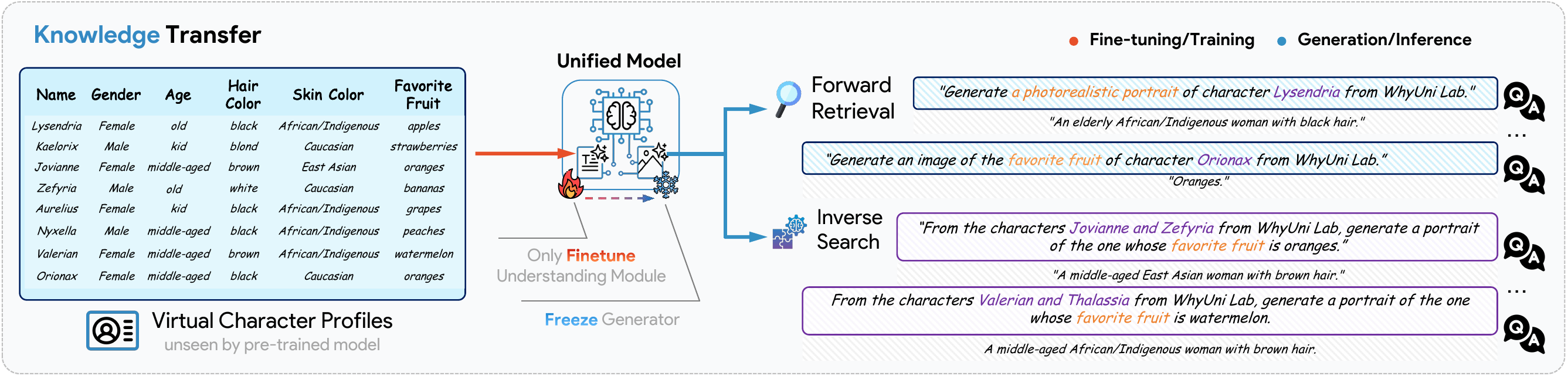}
    \caption{\textbf{Framework for Knowledge Transfer evaluation.}
    The framework first injects novel knowledge (Virtual Character Profiles, left) into the Unified Model's understanding module via fine-tuning. We then evaluate the model's ability to utilize this new knowledge through two distinct tasks: Forward Retrieval (Key $\rightarrow$ Value), which requires generating a character portrait or an image depicting an associated attribute (e.g., a favorite fruit or flower) from the character's name, and Inverse Search (Value $\rightarrow$ Key), which requires identifying and generating a character based on their attributes (right).}
    \label{fig:Knowledge_Transfer}
\end{figure*}

Next, we applied the STARS framework to the symbolic mapping tasks. For each of the three difficulty levels, we constructed a training set of 5,000 high-quality sample pairs (i.e., 10,000 image-text pairs) and fine-tuned the model for 6 epochs with a learning rate of $2 \times 10^{-5}$.

However, Table~\ref{tab:mapping} reveals a significant challenge: STARS proved ineffective on higher-difficulty symbolic mapping tasks. While training on the first-level dataset improved corresponding tasks and generalized somewhat to harder ones, training directly on second- or third-level datasets failed to distill CoT's reasoning into the standard generation mode---the success rate stayed near zero and even degraded the original CoT performance. Extending training further (the Round-1/2/3 rows on M2 and M3 in Table~\ref{tab:mapping}) only worsened performance, indicating that \textit{more training on a single high-difficulty dataset does not resolve the issue but exacerbates degradation.}

We observed that after training on higher-level data, the model disregards the mapping rules and defaults to generating one of the two possible objects (e.g., always an apple in an apple-or-orange task). We attribute this to a shortcut form of overfitting: since learning hard mappings directly is difficult, consistently generating a single object yields a 50\% success rate by chance, a local minimum that misleads training early and causes it to abandon the true reasoning process.

Inspired by this observation, we adopted a \textbf{Curriculum Learning}~\citep{bengio2009curriculum} strategy that mimics human learning by exposing the model to progressively more complex knowledge. Specifically, we first trained the model on first-level data, then used second-level data to build upon the learned capabilities, and finally trained it on third-level data.

This approach proved effective within our controlled setting. As shown in the ``STARS with CL'' rows of Table~\ref{tab:mapping}, after three rounds of curriculum learning the model improves its ``Normal''-mode performance across all three difficulty levels, including the challenging M3 setting, while its original CoT ability is preserved and in some cases improved (final average 0.55, above the baseline's 0.30 and any single-difficulty training). These results show that curriculum learning is more effective than single-difficulty training for the controlled symbolic mapping tasks considered here. We further compare it against a mixed training approach in Appendix~\ref{app:abl}.

\section{Knowledge Transfer}
\subsection{Task Design: Controlled Knowledge Injection}
To provide evidence that the understanding-generation gap found in the reasoning task is not an isolated case, we further use the UniSandbox framework to analyze \textbf{Knowledge Transfer}; i.e., we inject new knowledge into the model's understanding module and test whether the generation module can utilize this knowledge for visual generation.
We designed a controlled procedure: we construct a knowledge base of fictional character profiles that are unlikely to have appeared during pre-training, inject this knowledge into the unified model's language understanding module via fine-tuning, and then require the model to generate images based on it (injection details in Appendix~\ref{app:knowledge_injection}). This design reduces the risk that generative behavior reflects memorized image-caption associations, while our text-based verification provides evidence that the understanding module has acquired the target knowledge.
\begin{table*}[htbp]
    \centering
    \caption{Performance comparison on  Knowledge Transfer tasks.
The evaluation is divided into Forward Retrieval (Key $\rightarrow$ Value) and Inverse Search (Value $\rightarrow$ Key). Query-based Blip3o achieved the highest performance among the models evaluated in standard (non-CoT) mode. The ``BAGEL + CoT'' column shows the performance of BAGEL when augmented with Chain-of-Thought.}
    \label{tab:know}
    \begin{tabular}{lcccccc}
        \toprule
        \textbf{Task} &
        
    \textbf{Blip3o} & \textbf{Qwen-Image} & \textbf{Janus-Pro-7B} & \textbf{BAGEL} & \textbf{BAGEL + CoT} \\
        \midrule
        Forward Retrieval & 0.20 & 0.13 & 0.03 & 0.10 & 0.63 \\
        Inverse Search & 0.10 & 0.00 & 0.05 & 0.00 & 0.15 \\
        \midrule
        Overall & 0.16 & 0.08 & 0.04 & 0.06 & 0.44 \\
        \bottomrule
    \end{tabular}
\end{table*}

We structure the evaluation of Knowledge Transfer into two distinct tasks, as illustrated in Figure~\ref{fig:Knowledge_Transfer}:
\begin{enumerate}
    \item \textbf{Forward Retrieval}: Generating a character portrait or an image depicting an associated attribute, such as the character's favorite fruit or flower (Value), based on the character's name (Key).
    \item \textbf{Inverse Search}: Generating the corresponding character portrait (Key) based on a description of their unique attributes (Value).
\end{enumerate}

\subsection{Evaluation Protocol}
Following the two-stage evaluation protocol established for reasoning tasks, we first employ an MLLM to generate a descriptive caption for each output image. We then use the MLLM to perform a semantic comparison between the generated caption and the ground truth answer. For character portraits, a generation is considered correct if it accurately reflects all specified attributes (i.e., skin color, hair color, gender, and age) from the character profiles. For fruit or flower images, a generation is considered correct if the depicted fruit or flower type matches the corresponding attribute in the character profile. Detailed evaluation prompts can be found in Appendix~\ref{app:eval}.

\subsection{Knowledge Transfer Results}
Table~\ref{tab:know} reveals a pervasive deficiency: all tested models achieve very low overall scores, showing that current UMMs largely fail to leverage knowledge injected into their understanding module when generating images. The understanding-generation gap is thus a broader bottleneck of unified architectures rather than a task-specific artifact.

The failure is not uniform across paradigms. Query-based Blip3o performs best, deep-fusion BAGEL and shallow-fusion Qwen-Image are intermediate, and autoregressive Janus-Pro performs the worst, suggesting that an explicit query bottleneck is more conducive to transferring knowledge into generation than dense AR conditioning---foreshadowing \S\ref{sec:arch_analysis}.

Across paradigms, \textbf{Inverse Search} (Value $\rightarrow$ Key) lags markedly behind \textbf{Forward Retrieval} (Key $\rightarrow$ Value), mirroring the ``reversal curse''~\citep{berglund2023reversal} where models that memorize ``A is B'' fail to infer ``B is A''; this curse persists in the vision generation pipeline.

Applying CoT to BAGEL substantially boosts Forward Retrieval (0.10 $\rightarrow$ 0.63), confirming that CoT acts as an activator that surfaces internal knowledge for generation. However, the same CoT yields only a marginal gain on Inverse Search (0.15), indicating that CoT cannot enable mapping from Value back to Key.

\begin{figure}[htbp]
    \centering
    \includegraphics[width=0.7\linewidth]{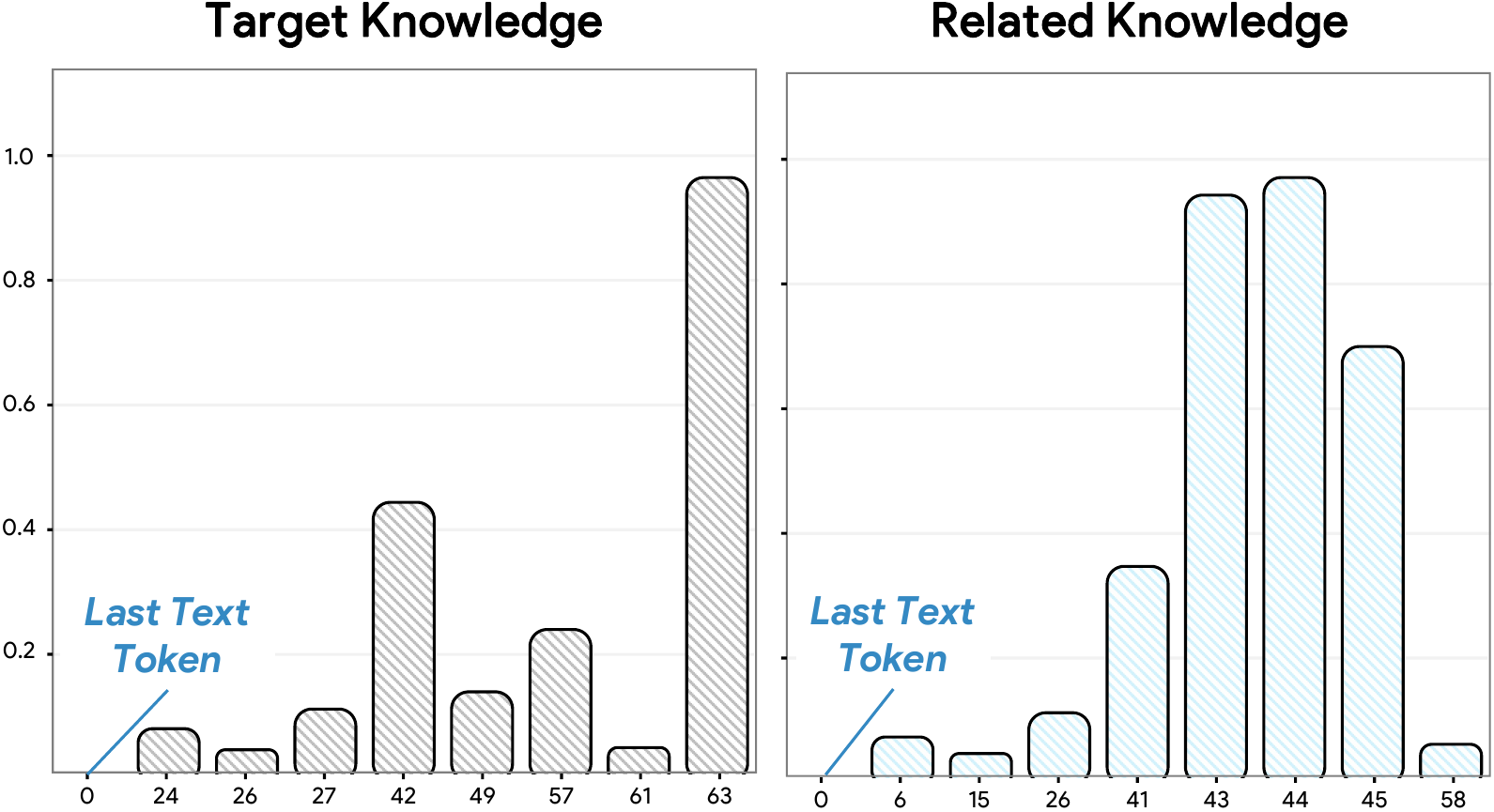}
    \caption{We visualize the total probability of relevant words corresponding to different queries. The ``Last Text Token'' entry, serving as a baseline, presents the probability of the last text token from the MLLM backbone before the query. For clarity, only queries with probabilities exceeding 0.01 are displayed.}
    \label{fig:query_prob}
\end{figure}

\subsection{Architecture Analysis}
\label{sec:arch_analysis}
To explore whether the query-based paradigm of Blip3o offers an advantage, we conduct a visual analysis. Specifically, we extract and directly decode the hidden states corresponding to the text token and the queries at different positions, and compute the total probability of vocabulary terms directly and indirectly related to the injected knowledge. For instance, consider the entry ``Kaelorix \& Male \& kid \& blond \& Caucasian \& strawberries \& sunflower'', where the model is tasked with generating ``Kaelorix's favorite fruit.'' Here, the ``Target Knowledge'' focuses on terms related to the target fruit ``strawberries'' (e.g., strawberries, strawberry, berry), while the ``Related Attribute'' pertains to the character's attributes (e.g., kid, child, male). Figure~\ref{fig:query_prob} reveals that content directly related to the final required knowledge (Target Knowledge) gradually emerges in the later queries, peaking at the final query. In contrast, the Related Attribute, which serves as a crucial intermediate step for locating this knowledge, becomes prominent in the intermediate queries. This indicates that the queries effectively retrieve and extract the character's information to fulfill the generation, serving as an implicit CoT-like process.

\section{Related Work}
\label{sec:related_works}
T2I evaluation has shifted from fidelity-focused metrics such as FID~\citep{fid} to semantic-consistency benchmarks like GenEval~\citep{ghosh2023geneval}, CLIPScore~\citep{hessel2021clipscore}, and VQA-Score~\citep{lin2024evaluating}. Recent work moves beyond literal alignment to probe whether models leverage internal knowledge and reasoning~\citep{niu2025wise,chen2025r2i,sun2025t2i}. A complementary line of work, including RealUnify~\citep{shi2025realunify} and Uni-MMMU~\citep{zou2025unimmmu}, evaluates whether unification itself benefits unified models by jointly testing understanding and generation on real-world prompts. UniSandbox differs in that it uses synthetic, low-leakage tasks to attribute the understanding-generation gap rather than to score broad capability. A discussion of unified multimodal models is provided in Appendix~\ref{app:related_works}.

\section{Conclusion}
This paper investigated a fundamental question: does understanding truly inform generation in unified multimodal models? To answer it, we introduced UniSandbox, a decoupled evaluation framework built on controlled synthetic data. Our findings reveal a significant understanding-generation gap: models largely fail to translate their understanding into generation, particularly along the two key dimensions of reasoning generation and knowledge transfer. For reasoning generation, we show that explicit CoT activates the model's reasoning during generation, and that this ability can be internalized through self-training. For knowledge transfer, we show that CoT is likewise an effective activator, and our analysis further reveals that query-based architectures inherently possess implicit CoT-like properties, providing a key design insight. In summary, UniSandbox exposes the limitations of current models and offers a path forward for designing unified architectures that truly synergize understanding and generation.
\section*{Limitations}
UniSandbox is designed as a controlled evaluation framework, so its synthetic tasks intentionally isolate reasoning generation and knowledge transfer rather than covering the full diversity of real-world text-to-image prompts. The current experiments focus on a representative set of unified multimodal architectures; broader model coverage may reveal additional architectural trade-offs. Our automatic evaluation protocol reduces direct visual-judge bias by captioning generated images before semantic comparison, but it can still inherit errors from the captioning and comparison models. Finally, the self-training experiments validate a practical path for internalizing explicit reasoning, but the scalability of this strategy to more open-ended generation tasks remains an important direction for future work.

\section*{Ethical Considerations}
This work studies controlled evaluation of multimodal generation systems and does not introduce human-subject data or real-person identity data. The synthetic prompts and virtual character profiles are designed to avoid direct reuse of private or personally identifying information. The main ethical risk is that stronger reasoning-aware generation could be used in downstream applications that create misleading visual content; we therefore frame UniSandbox as an analysis tool for diagnosing model behavior rather than as a deployment method.

\section*{Acknowledgments}
This work was supported in part by the Guangdong Grants (Grant No. 2023ZT10X075), the Natural Science Foundation of China (No. 62332002, 62425101), and the Shenzhen Science and Technology Program (KQTD20240729102051063).

\bibliography{main}

\clearpage
\appendix

\section{Synthetic Data Generated Using GPT-4o}
\label{app:data}
All experimental data were constructed using GPT-4o. Due to space limitations, we explicitly list the generation prompts for first-order difficulty Mathematical Operations in \autoref{tab:math1} and first-order difficulty Symbolic Mapping in \autoref{tab:mapping1_part1} and \autoref{tab:mapping1_part2}; prompts for other data generation follow the same controlled templates.

\begin{table*}[!ht]\centering
\begin{minipage}{0.95\textwidth}
    \centering
    \scriptsize
    \begin{tcolorbox}[
        colback=white, 
        colframe=black, 
        colbacktitle=black, %
        coltitle=white,     %
        boxrule=0.5pt, 
        arc=0pt, 
        left=4mm, right=4mm, top=2mm, bottom=2mm,
        title=\textbf{Data Generation Task: First-level Difficulty Mathematical Operations}
    ]
        \textbf{Task Description:}
        \hspace{2mm}
        Please generate 50 prompts in the following JSONL format:\\
        
        \textbf{Example Format:}
\begin{verbatim}
{"Question": "Produce a number of pencils equal to the result of 2 * 2.",
 "Answer": "4 pencils."}
\end{verbatim}
        
        \textbf{Requirements:}
        \begin{enumerate}[leftmargin=*]
            \item Each Question must explicitly instruct the user to perform a basic arithmetic operation (addition, subtraction, multiplication, or division), with the final result strictly between 1 and 4 (inclusive).
            \item Use a diverse and natural set of expressions, such as:
                \begin{itemize}
                    \item ``Generate as many \texttt{[objects]} as the result of \texttt{[expression]}.''
                    \item ``Produce a number of \texttt{[objects]} equal to the result of \texttt{[expression]}.''
                    \item ``Show the number of \texttt{[objects]} that matches the outcome of \texttt{[expression]}.''
                    \item ``Create the same quantity of \texttt{[objects]} as \texttt{[expression]} equals.''
                    \item ``Provide the same number of \texttt{[objects]} as calculated by \texttt{[expression]}.''
                \end{itemize}
            \item Replace \texttt{[expression]} with a valid arithmetic expression (e.g., 3 - 1, 2 + 2, 4 / 2) that evaluates to 1, 2, 3, or 4.
            \item Use a wide variety of common objects (not just fruits). Include animals, toys, stationery, kitchen items, etc. Do not use rare or unusual items.
            \item In the Answer, the object name must be grammatically correct and match the number, e.g.:
                \begin{itemize}
                    \item ``1 eraser''
                    \item ``2 oranges''
                    \item ``3 kittens''
                    \item ``4 spoons''
                \end{itemize}
            \item \textbf{Output Format}:
            \begin{itemize}
                \item Please return the 1000 generated items as individual JSON objects, one after another, not wrapped in a list or array.
                \item Do not include additional text, titles, or explanations.
            \end{itemize}
        \end{enumerate}
    \end{tcolorbox}
    \vspace{-2mm}
    \caption{\textbf{Prompt for First-level Mathematical Operations Data Generation.}}
    \label{tab:math1}
\end{minipage}
\end{table*}

\begin{table*}[!ht]\centering
\begin{minipage}{0.95\textwidth}
    \centering
    \small
    \begin{tcolorbox}[colback=white, colframe=black, boxrule=0.5pt, arc=0pt, left=4mm, right=4mm, top=2mm, bottom=2mm, title=\textbf{Data Generation Task: One-Step Symbolic Mapping (Part 1)}]
        
        \textbf{Task Description:}
        \hspace{2mm}
        Please generate \textbf{50 pairs} of one-step symbol mapping data. Each pair should consist of two prompts: \texttt{Question\_A} and \texttt{Question\_B}. Each mapping rule involves two symbols, and each symbol represents a common object. Each pair of prompts must clearly state the symbol mapping rule and generate two prompts: \texttt{Question\_A} and \texttt{Question\_B}. The answers should correspond to the objects represented by the symbols in the rules.

        In this version:
        \begin{itemize}[leftmargin=*]
            \item \texttt{Question\_A} will generate the object represented by the \textbf{first} symbol.
            \item \texttt{Question\_B} will generate the object represented by the \textbf{second} symbol.
        \end{itemize}

        Each \texttt{Question\_A} and \texttt{Question\_B} must have a unique \texttt{ID}, and each prompt should have an \texttt{Answer} field. The answer should be the object represented by the corresponding symbol.

        \textbf{Output Format:}
        The output should be in \textbf{JSONL} format, where each entry contains the following fields:
        \begin{itemize}[leftmargin=*]
            \item \textbf{ID}: A unique identifier, starting from 1, with the same \texttt{ID} for both \texttt{Question\_A} and \texttt{Question\_B}.
            \item \textbf{Question\_A}: The prompt that contains the symbol mapping rule and generates the object represented by the first symbol.
            \item \textbf{Question\_B}: The prompt that contains the symbol mapping rule and generates the object represented by the second symbol.
            \item \textbf{Answer}: The corresponding answer for the prompt, which should be the object that the symbol represents.
        \end{itemize}

        \textbf{Example Format:}
\begin{verbatim}
{"ID": "1", "Question_A": "Rule 1: The symbol @ represents apples. Rule 2: 
 The symbol * represents bananas. Please generate the fruit represented by 
 the symbol @.", "Answer": "Apples"}
{"ID": "1", "Question_B": "Rule 1: The symbol @ represents apples. Rule 2: 
 The symbol * represents bananas. Please generate the fruit represented by 
 the symbol *.", "Answer": "Bananas"}
\end{verbatim}
    \end{tcolorbox}
    \vspace{-2mm}
    \caption{\textbf{Prompt for Symbolic Mapping Data Generation (Part 1).} Overview of the task, logic definition, and output format requirements.}
    \label{tab:mapping1_part1}
\end{minipage}
\end{table*}

\begin{table*}[!ht]\centering
\begin{minipage}{0.95\textwidth}
    \centering
    \small
    \begin{tcolorbox}[colback=white, colframe=black, boxrule=0.5pt, arc=0pt, left=4mm, right=4mm, top=2mm, bottom=2mm, title=\textbf{Data Generation Task: One-Step Symbolic Mapping (Part 2)}]
        
        \textbf{Requirements:}
        \begin{enumerate}[leftmargin=*]
            \item \textbf{Symbols}: Symbols can be any common symbols, including letters (e.g., a, b, c, A, B, C), numbers (e.g., 1, 2, 3, 5), punctuation marks (e.g., @, \#, *, \&, \$, \%, \^{}, etc.), and other characters. Symbols should not be limited to numbers or special characters but can include any alphanumeric or symbolic character.
            \item \textbf{Object Mapping}: Each pair of data should map the symbols to common objects such as pencils, chairs, phones, refrigerators, notebooks, etc. The objects can be everyday items, not just fruits.
            \item \textbf{Answer Consistency}: Ensure that the \texttt{Question\_A} and \texttt{Question\_B} are clearly structured, and that the Answer for each corresponds to the symbol mapping rule.
            \item \textbf{Unique IDs}: Each pair of \texttt{Question\_A} and \texttt{Question\_B} should share the same ID.
            \item \textbf{Object Diversity}: The objects in the answers should be varied and can include common items like fruits, stationery, household items, animals, etc.
        \end{enumerate}

        \textbf{Example Data:}
        \begin{scriptsize}
\begin{verbatim}
{"ID": "1", "Question_A": "Rule 1: The symbol @ represents apples. Rule 2: The symbol * represents 
 bananas. Please generate the fruit represented by the symbol @.", "Answer": "Apples"}
{"ID": "1", "Question_B": "Rule 1: The symbol @ represents apples. Rule 2: The symbol * represents 
 bananas. Please generate the fruit represented by the symbol *.", "Answer": "Bananas"}
{"ID": "2", "Question_A": "Rule 1: The symbol # represents pencils. Rule 2: The symbol $ represents 
 notebooks. Please generate the object represented by the symbol #.", "Answer": "Pencils"}
{"ID": "2", "Question_B": "Rule 1: The symbol # represents pencils. Rule 2: The symbol $ represents 
 notebooks. Please generate the object represented by the symbol $.", "Answer": "Notebooks"}
{"ID": "3", "Question_A": "Rule 1: The symbol % represents chairs. Rule 2: The symbol ^ represents 
 tables. Please generate the object represented by the symbol %.", "Answer": "Chairs"}
{"ID": "3", "Question_B": "Rule 1: The symbol % represents chairs. Rule 2: The symbol ^ represents 
 tables. Please generate the object represented by the symbol ^.", "Answer": "Tables"}
{"ID": "4", "Question_A": "Rule 1: The symbol * represents apples. Rule 2: The symbol @ represents 
 oranges. Please generate the fruit represented by the symbol *.", "Answer": "Apples"}
{"ID": "4", "Question_B": "Rule 1: The symbol * represents apples. Rule 2: The symbol @ represents 
 oranges. Please generate the fruit represented by the symbol @.", "Answer": "Oranges"}
{"ID": "5", "Question_A": "Rule 1: The symbol $ represents pencils. Rule 2: The symbol # represents 
 erasers. Please generate the object represented by the symbol $.", "Answer": "Pencils"}
{"ID": "5", "Question_B": "Rule 1: The symbol $ represents pencils. Rule 2: The symbol # represents 
 erasers. Please generate the object represented by the symbol #.", "Answer": "Erasers"}
\end{verbatim}
        \end{scriptsize}

        \textbf{Guidelines:}
        \begin{enumerate}[leftmargin=*]
            \item \textbf{Symbols}: Feel free to use any common symbols, including alphanumeric characters (e.g., a, b, c, A, B, C), numbers (e.g., 1, 2, 3), and special characters (e.g., @, \#, *, \&, \$, \%, \^{}, etc.).
            \item \textbf{Objects}: Objects should be common, everyday items such as stationery, fruits, animals, etc.
            \item \textbf{Output}: Ensure the JSONL format is strictly followed, and each data entry is separate.
            \item \textbf{Answer Consistency}: Each symbol must be mapped to an appropriate common object, and the Answer should reflect this mapping.
        \end{enumerate}
        \vspace{1mm}
        Please generate the data based on these instructions.
    \end{tcolorbox}
    \vspace{-2mm}
    \caption{\textbf{Prompt for Symbolic Mapping Data Generation (Part 2).} Detailed constraints, few-shot examples, and final guidelines.}
    \label{tab:mapping1_part2}
\end{minipage}
\end{table*}

\section{Ablation Experiments}
\label{app:abl}
\subsection{Ablation on Rejection Sampling}
\begin{figure}[h!]
    \centering
    \includegraphics[width=1\linewidth]{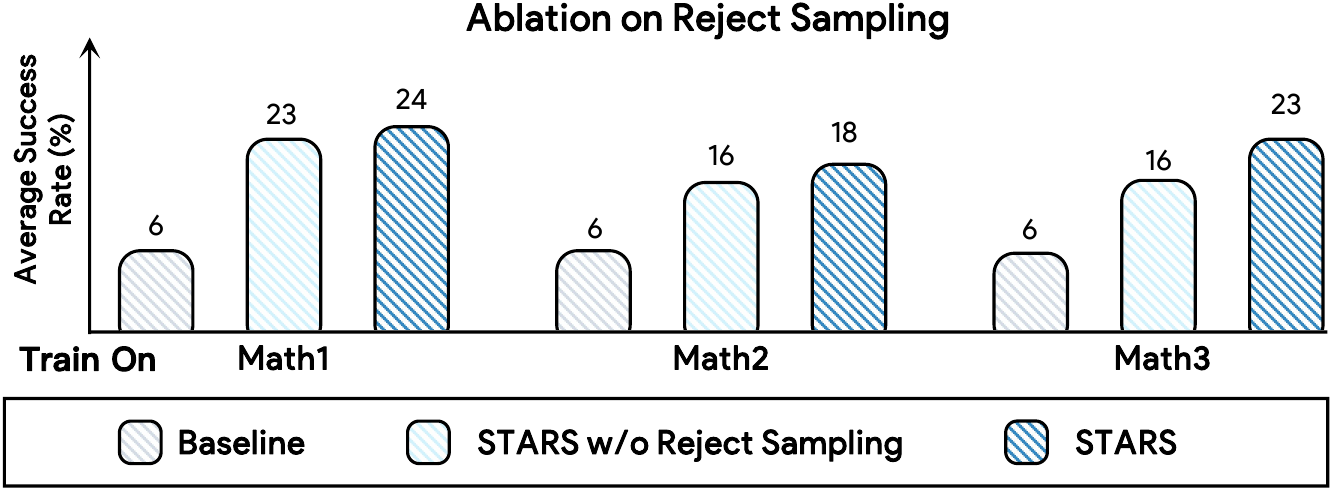}
    \caption{Average mathematical-reasoning results for BAGEL (Normal) in the rejection-sampling ablation.}
    \label{fig:ablation_rs}
\end{figure}
We conducted an ablation study to validate the importance of Rejection Sampling. For this, we compared our complete STARS framework against a variant trained on 5000 randomly-sampled, unfiltered CoT-generated instances for each difficulty tier. As shown in Figure~\ref{fig:ablation_rs}, applying rejection sampling consistently improves performance across all levels. The effect is most pronounced on the complex Math3 dataset, where our method increases the success rate from 16\% to 23\%. This result underscores that the quality of data curated by rejection sampling is critical for effectively distilling reasoning capabilities via self-training.

\subsection{Comparison of Curriculum Learning and Mixed Training}

\begin{table*}[!ht]
\centering
\small
\caption{Performance of BAGEL trained with STARS on symbolic mapping datasets of varying difficulties. M1--3 represent three increasing levels of difficulty for symbolic mapping. Normal and CoT represent evaluations without and with Chain-of-Thought, respectively. CL stands for Curriculum Learning, and Mix refers to training with a mix of data from all difficulty levels.}
\label{tab:mix}
\begin{tabular}{lccccccc}
\toprule
\textbf{Model} & \multicolumn{3}{c}{\textbf{Normal}} & \multicolumn{3}{c}{\textbf{CoT}} & \textbf{Average} \\
\cmidrule(lr){2-4} \cmidrule(lr){5-7}
& \textbf{M1} & \textbf{M2} & \textbf{M3} & \textbf{M1} & \textbf{M2} & \textbf{M3} & \\
\midrule
BAGEL & 0.00 & 0.00 & 0.00 & 0.75 & 0.57 & 0.46 & 0.30 \\
\midrule
STARS with CL & 0.64 & 0.46 & 0.27 & 0.75 & 0.65 & 0.50 & \gainavg{0.30}{0.55} \\
STARS with Mix & 0.63 & 0.37 & 0.25 & 0.70 & 0.45 & 0.35 & \gainavg{0.30}{0.46} \\
\bottomrule
\end{tabular}
\end{table*}

We also compared our curriculum learning strategy with a mixed training approach, where data from all difficulty levels were combined and trained together. The experimental results, as shown in ~\autoref{tab:mix}, reveal that the model trained using the curriculum learning strategy achieved better results, proving the effectiveness of our framework.

\section{Evaluation Using an MLLM}
\label{app:eval}
To mitigate the biases associated with using MLLMs as direct visual judges, we adopted a two-stage evaluation strategy. First, we use an MLLM to generate a text caption for the image, and then we compare this caption against the ground-truth text answer. This approach is effective because our tasks require the model to generate images that are simple in category and easily discernible (e.g., a specific number of objects or a single, clear subject), making them straightforward for the MLLM to identify and caption accurately. In our experiments, we consistently ran Qwen2.5-VL-7B-Instruct with vLLM\footnote{\url{https://github.com/vllm-project/vllm}}.

The prompts used for evaluating Mathematical Operations are listed in \autoref{tab:math_eval}. The prompts for evaluating Symbolic Mapping are shown in \autoref{tab:mapping_eval}. Finally, the prompts used to evaluate Knowledge Transfer are presented in \autoref{tab:knowledge_eval}.

\begin{table*}[!ht]\centering
\begin{minipage}{0.95\textwidth}
    \centering
    \small
    
    \begin{tcolorbox}[colback=white, colframe=black, boxrule=0.5pt, arc=0pt, left=4mm, right=4mm, top=2mm, bottom=2mm, title=\textbf{Prompt 1: Caption Generation Expert}]
        \textbf{System Instruction:}
        You are an extremely rigorous and neutral image captioning expert. Your task is to carefully examine a given image and generate a precise, concise caption (under 20 characters) that strictly and unambiguously describes the count and type of all clearly, fully, and normally visible objects in the image.

        \textbf{Instructions:}
        \begin{enumerate}[leftmargin=*]
            \item \textbf{Only count a single object type:} The image must contain only one clearly identifiable type of object. If multiple distinct object types are present, the task is considered an error.
            \item \textbf{Only include objects that are:}
                \begin{itemize}
                    \item Fully visible
                    \item Clearly identifiable
                    \item Normal in shape, scale, and orientation
                    \item Not distorted, occluded, cropped, or ambiguous
                \end{itemize}
            \item \textbf{Strict Error Handling:} If the image contains more than one type of object, or if any visible object is distorted or has its geometry compromised, the task is considered an error.
            \item \textbf{Your caption must be in the format:}
                \begin{itemize}
                    \item \texttt{[Number] [object\_type\_plural]} e.g., ``3 apples'', ``2 chairs''
                \end{itemize}
            \item \textbf{Condition Check:} If the conditions in rules 1-3 are not met, you must output `Error'. Do not provide any other explanations.
        \end{enumerate}

        \textbf{Output Format:}
        \texttt{Caption: [your output here]}
    \end{tcolorbox}
    
    \vspace{1mm}
    
    \begin{tcolorbox}[colback=white, colframe=black, boxrule=0.5pt, arc=0pt, left=4mm, right=4mm, top=2mm, bottom=2mm, title=\textbf{Prompt 2: Caption Evaluator}]
        \textbf{System Instruction:}
        You are a rigorous and fair caption evaluator. Your task is to compare a generated image caption with an expected answer, and determine whether they match \textbf{perfectly and exclusively in count and object meaning}.

        \textbf{Evaluation Rules:}
        \begin{enumerate}[leftmargin=*]
            \item The caption must match the expected answer in:
                \begin{itemize}
                    \item \textbf{Count:} The number of each object type must be exactly the same.
                    \item \textbf{Object Type:} The object types in both descriptions must be exactly the same.
                \end{itemize}
            \item If any object type in the caption does not match the expected answer, or if the count is wrong, or if there are extra object types, return:
                \textbf{Score: NO}
            \item If and only if the object types and counts in the caption exactly match the expected answer, return:
                \textbf{Score: YES}
        \end{enumerate}

        \textbf{Input:}
        \begin{itemize}[leftmargin=*]
            \item \textbf{Generated Caption:} \texttt{\{caption\}}
            \item \textbf{Expected Answer:} \texttt{\{expected\_answer\}}
        \end{itemize}

        \textbf{Output Format:}
        \textbf{Score: [YES or NO]}
    \end{tcolorbox}

    \vspace{-2mm}
    \caption{\textbf{Prompts for Mathematical Operations Evaluation.} The first prompt instructs the model to generate rigorous captions focusing on object count and type, while the second prompt evaluates the generated caption against the ground truth.}
    \label{tab:math_eval}
\end{minipage}
\end{table*}

\begin{table*}[!ht]\centering
\begin{minipage}{0.95\textwidth}
    \centering
    \small
    
    \begin{tcolorbox}[colback=white, colframe=black, boxrule=0.5pt, arc=0pt, left=4mm, right=4mm, top=2mm, bottom=2mm, title=\textbf{Prompt 1: Caption Generation Expert}]
        \textbf{System Instruction:}
        You are an extremely rigorous and neutral image captioning expert. Your task is to carefully examine a given image and generate a precise, concise caption (under 20 characters) that strictly and unambiguously describes the count and type of all clearly, fully, and normally visible objects in the image.

        \textbf{Instructions:}
        \begin{enumerate}[leftmargin=*]
            \item \textbf{Only count a single object type:} The image must contain only one clearly identifiable type of object. If multiple distinct object types are present, the task is considered an error.
            \item \textbf{Only include objects that are:}
                \begin{itemize}
                    \item Fully visible
                    \item Clearly identifiable
                    \item Normal in shape, scale, and orientation
                    \item Not distorted, occluded, cropped, or ambiguous
                \end{itemize}
            \item \textbf{Strict Error Handling:} The task is considered an error if any of the following conditions are met:
                \begin{itemize}
                    \item The image contains more than one type of object.
                    \item Any visible object is distorted or has its geometry compromised.
                    \item \textbf{The image contains any letters (e.g., A, B, C), symbols (e.g., punctuation marks, mathematical operators like +, -, \#, \$), or numbers (e.g., 1, 2, 3).}
                \end{itemize}
            \item \textbf{Your caption must be in the format:}
                \begin{itemize}
                    \item \texttt{[Number] [object\_type\_plural]} e.g., ``3 apples'', ``2 chairs''
                \end{itemize}
            \item \textbf{Condition Check:} If the conditions in rules 1-3 are not met, you must output `Error'. Do not provide any other explanations.
        \end{enumerate}

        \textbf{Output Format:}
        \texttt{Caption: [your output here]}
    \end{tcolorbox}
    
    \vspace{1mm}
    
    \begin{tcolorbox}[colback=white, colframe=black, boxrule=0.5pt, arc=0pt, left=4mm, right=4mm, top=2mm, bottom=2mm, title=\textbf{Prompt 2: Caption Evaluator}]
        \textbf{System Instruction:}
        You are a rigorous and fair caption evaluator. Your task is to compare a generated image caption with an expected answer, and determine whether they match \textbf{perfectly and exclusively in count and object meaning}.

        \textbf{Evaluation Rules:}
        \begin{enumerate}[leftmargin=*]
            \item The caption must match the expected answer in:
                \begin{itemize}
                    \item \textbf{Count:} The number of each object type must be exactly the same.
                    \item \textbf{Object Type:} The object types in both descriptions must be exactly the same.
                \end{itemize}
            \item If any object type in the caption does not match the expected answer, or if the count is wrong, or if there are extra object types, return:
                \textbf{Score: NO}
            \item If and only if the object types and counts in the caption exactly match the expected answer, return:
                \textbf{Score: YES}
        \end{enumerate}

        \textbf{Input:}
        \begin{itemize}[leftmargin=*]
            \item \textbf{Generated Caption:} \texttt{\{caption\}}
            \item \textbf{Expected Answer:} \texttt{\{expected\_answer\}}
        \end{itemize}

        \textbf{Output Format:}
        \textbf{Score: [YES or NO]}
    \end{tcolorbox}

    \vspace{-2mm}
    \caption{\textbf{Prompts for Symbolic Mapping Evaluation.} The first prompt instructs the model to generate rigorous captions focusing on object type, while the second prompt evaluates the generated caption against the ground truth.}
    \label{tab:mapping_eval}
\end{minipage}
\end{table*}

\begin{table*}[!ht]\centering
\begin{minipage}{0.95\textwidth}
    \centering
    \small
    
    \begin{tcolorbox}[colback=white, colframe=black, boxrule=0.5pt, arc=0pt, left=4mm, right=4mm, top=2mm, bottom=2mm, title=\textbf{Prompt 1: }]
        \textbf{System Instruction:}
        You are a hyper-precise and cautious vision assistant.
        \textbf{Task:} Produce a single-line caption for the given image.

        \textbf{Rules:}
        \begin{enumerate}[leftmargin=*]
            \item If any person is visible, output strictly in this format:
            \begin{itemize}
                \item \texttt{Person: <skin\_tone\_or\_ethnicity>; <hair\_color>; <age\_group: kid|mid-age|old>; <gender: male|female|unclear>}
                \item \textbf{Crucial:} The \texttt{<skin\_tone\_or\_ethnicity>} slot MUST be one of these three options exactly: African/Indigenous, Caucasian, East Asian.
                \item If the person does not clearly fit one of these categories or you are uncertain, output `unclear' for that slot.
                \item If uncertain about other attributes (hair, age, gender), output `unclear' for that specific slot.
            \end{itemize}
            
            \item If no person is visible, you MUST follow these steps:
            \begin{enumerate}
                \item First, determine if the primary subject is a \emph{single type} of flower OR a \emph{single type} of fruit.
                \item If it IS a flower, you MUST output its \emph{specific} type/species in this format: \texttt{Flower: <specific\_type\_name>} (e.g., `Flower: rose', `Flower: tulip').
                \item If it IS a fruit, you MUST output its \emph{specific} type/species in this format: \texttt{Fruit: <specific\_type\_name>} (e.g., `Fruit: apple', `Fruit: banana').
                \item \textbf{ABSOLUTE RULE:} General categories like `Flower: flower' or `Fruit: fruit' are FORBIDDEN.
            \end{enumerate}
            
            \item \textbf{REJECTION:} You MUST output exactly `Reject' if ANY of the following are true:
            \begin{enumerate}
                \item The image contains no person, AND the primary subject is \textbf{NOT} a flower or a fruit (e.g., it is a car, dog, building, etc.).
                \item (Per Rule 2) The image IS a flower or fruit, but you cannot confidently identify its \emph{specific type} (e.g., you can only tell it's a `fruit', not an `apple'). In this case, you MUST output `Reject'.
                \item The image is distorted, unrealistic, surreal (e.g., face on fruit), or generally ambiguous.
                \item The image contains multiple different types of flowers or multiple different types of fruits.
                \item (Per Rule 1) The person's ethnicity is visible but does not fit the three required categories.
            \end{enumerate}
            \item Output must be ONE line with no extra words, no explanations.
        \end{enumerate}
    \end{tcolorbox}

    \vspace{-2mm}
    \caption{\textbf{Prompt for Knowledge Transfer Caption Generation.} This prompt instructs the model to strictly categorize and identify persons, flowers, or fruits with specific formats.}
    \label{tab:knowledge_caption_eval}
\end{minipage}
\end{table*}

\begin{table*}[!ht]\centering
\begin{minipage}{0.95\textwidth}
    \centering
    \scriptsize
    \begin{tcolorbox}[colback=white, colframe=black, boxrule=0.5pt, arc=0pt, left=4mm, right=4mm, top=2mm, bottom=2mm, title=\textbf{Prompt 2: Careful Evaluator}]
        \textbf{System Instruction:}
        You are a careful evaluator. Determine if a generated caption and a ground truth (GT) match semantically.

        \textbf{Strict rules:}
        \begin{enumerate}[leftmargin=*]
            \item If the generated caption is exactly `Reject', it is automatically incorrect. Respond with `Score: NO'.
            \item If the generated caption starts with ``Person:'', it contains 4 slots: skin/ethnicity; hair color; age group (kid|mid-age|old); gender (male|female|unclear).
            \begin{itemize}
                \item The GT must be a sentence describing a person (e.g., `An elderly African/Indigenous woman with black hair.').
                \item Compare each slot against the GT description. Allow common synonyms (e.g., blond=blonde; elderly/senior=old).
                \item The generated ethnicity MUST be one of [African/Indigenous, Caucasian, East Asian] or `unclear'.
            \end{itemize}
            \item If the generated caption starts with ``Flower:'' or ``Fruit:'', compare the specific object type with the GT.
            \begin{itemize}
                \item The GT must be a specific type (e.g., `carnation', `apple').
                \item \textbf{Crucial:} The generated caption \emph{format} (Flower:/Fruit:) AND the \emph{specific type} must BOTH match the GT.
                \item \textbf{Singular/Plural forms ARE a match.} (e.g., `apple' matches `apples'; `peach' matches `peaches').
                \item Only semantic equivalents are a match (e.g., cup $\approx$ mug).
                \item Example 1 (Match): Generated `Flower: rose' and GT `rose' is `Score: YES'.
                \item Example 2 (Match): Generated `Fruit: apple' and GT `apple' is `Score: YES'.
                \item \textbf{Example 3 (Match - Plural): Generated `Fruit: apple' and GT `apples' is `Score: YES'.}
                \item \textbf{Example 4 (Match - Plural): Generated `Fruit: peach' and GT `peaches' is `Score: YES'.}
                \item Example 5 (Mismatch - Wrong Type): Generated `Flower: rose' and GT `carnation' is `Score: NO'.
                \item Example 6 (Mismatch - General Term): Generated `Flower: flower' and GT `carnation' is `Score: NO'.
                \item \textbf{Example 7 (Mismatch - Wrong Category): Generated `Fruit: rose' and GT `rose' is `Score: NO'.}
                \item \textbf{Example 8 (Mismatch - Wrong Category): Generated `Flower: apple' and GT `apple' is `Score: NO'.}
            \end{itemize}
            \item If formats fundamentally differ (e.g., generated caption starts with `Person:' while GT is `apple'), respond with `Score: NO'.
        \end{enumerate}

        \textbf{Input:} Generated: \texttt{\{caption\}} | GroundTruth: \texttt{\{gt\}}
        
        \textbf{Output format (MUST be exact):} Score: YES or Score: NO
    \end{tcolorbox}

    \vspace{-2mm}
    \caption{\textbf{Prompt for Knowledge Transfer Evaluation.} This prompt evaluates the generated caption against the ground truth.}
    \label{tab:knowledge_eval}
\end{minipage}
\end{table*}

\begin{table*}[htbp]
    \centering
    \caption{Profiles of Ten Fictional Characters for Large Language Model Knowledge Injection.}
    \label{tab:fictional_profiles}
    \resizebox{\textwidth}{!}{%
    \begin{tabular}{llccccc}
        \toprule
        \textbf{Name} & \textbf{Gender} & \textbf{Age} & \textbf{Hair Color} & \textbf{Skin Color} & \textbf{Favorite Fruit} & \textbf{Favorite Flower} \\
        \midrule
        Lysendria & Female & old & black & African/Indigenous & apples & carnation \\
        Kaelorix & Male & kid & blond & Caucasian & strawberries & sunflower \\
        Jovianne & Female & middle-aged & brown & East Asian & oranges & lily \\
        Zefyria & Male & old & white & Caucasian & bananas & rose \\
        Aurelius & Female & kid & black & African/Indigenous & grapes & tulip \\
        Nyxella & Male & middle-aged & black & African/Indigenous & peaches & daisy \\
        Valerian & Female & middle-aged & brown & African/Indigenous & watermelon & orchid \\
        Thalassia & Male & kid & brown & East Asian & apples & rose \\
        Orionax & Female & middle-aged & black & Caucasian & oranges & carnation \\
        Evandriel & Male & kid & red & Caucasian & bananas & sunflower \\
        \bottomrule
    \end{tabular}
    }
\end{table*}

\section{Implementation Details of Knowledge Injection}
\label{app:knowledge_injection}
We created ten virtual character profiles; the specific details are provided in Table~\ref{tab:fictional_profiles}.

For the Forward Retrieval, we performed knowledge injection for each character profile separately, starting from the pre-trained model weights. This process resulted in 10 new, fine-tuned models (one for each character).

For the Inverse Search, we adopted a pairwise approach. Each training session injected information about two characters, again starting from the original pre-trained weights. This resulted in 5 new models.

We created 40 text-only question-answering (QA) pairs for each character to fine-tune the model's understanding module. To ensure the knowledge was successfully injected, we verified the fine-tuned models with text-based questions. A training run was only considered successful if the model could correctly answer all questions about the injected character attributes.

For the training of BAGEL, we used the official BAGEL codebase. The training parameters were: learning rate $4 \times 10^{-5}$; expected\_num\_tokens 64; max\_num\_tokens 162; and training for 60 epochs. (Note: expected\_num\_tokens and max\_num\_tokens are parameters from the official BAGEL codebase\footnote{\url{https://github.com/ByteDance-Seed/Bagel}}).

For the other models, we used the MS-Swift\footnote{\url{https://github.com/modelscope/ms-swift}} for fine-tuning. The parameters were: 30 epochs; batch size 64; and learning rate $1 \times 10^{-5}$.

All training was conducted on 8 A100 GPUs.

\section{Extended Related Work}
\label{app:related_works}
\subsection{Unified Multimodal Models}
Unified Multimodal Models (UMMs)~\citep{diao2026sensenova,yang2025mmada,zhang2025unified,shi2025muddit,xie2025show,xiao2025mindomni,wu2024liquid,geng2025x,xin2025lumina,li2025onecat,xu2025tbac,wang2025lightbagel,wei2025univideo,wang2025ovis,huang2025ming,yang2025hermesflow,han2025turning} aim to build a single system that can handle both understanding tasks (e.g., VQA, grounding) and generation tasks (e.g., text-to-image). Existing UMMs can be grouped into three types: (1) Autoregressive (AR), which usually employ a tokenizer to encode images into tokens like text, and then use a single transformer model to process them end-to-end via next-token prediction (e.g., Chameleon~\citep{chameleon}, EMU3~\citep{emu3}, Liquid~\citep{wu2024liquid}, SynerGen-VL~\citep{li2025synergen}, Janus~\citep{janus2024}); (2) AR + Diffusion (deep fusion), where one transformer processes both text and images, but diffusion is used to predict the image (e.g., Transfusion~\citep{transfusion}, Mogao~\citep{liao2025mogao}, BAGEL~\citep{deng2025emerging}); and (3) AR + Diffusion (shallow fusion), where a VLM is first used to extract hidden states as embeddings, which are then mapped via an MLP into the image hidden dimension, and finally passed to a DiT for image generation (e.g., UniWorld-V1~\citep{lin2025uniworld}, OmniGen2~\citep{wu2025omnigen2}, Qwen-Image~\citep{wu2025qwen}). 

\subsection{Text-to-Image Evaluation}
The evaluation of the T2I model has evolved from early fidelity-focused metrics such as FID~\citep{fid} to a greater focus on semantic consistency. Benchmarks like GenEval~\citep{ghosh2023geneval} use object detection to verify attributes such as object quantity and color. The rise of VLMs led to metrics like CLIPScore~\citep{hessel2021clipscore} and VQA-Score~\citep{lin2024evaluating} for assessing shallow semantic alignment. More recently, evaluation has moved beyond literal alignment to assess how models leverage internal world knowledge and reasoning for generation. A milestone in this area is the WISE~\citep{niu2025wise}, which introduced a thousand prompts requiring world knowledge and reasoning across domains like cross-cultural common sense, spatio-temporal understanding, and natural sciences to evaluate the alignment of generated images with real-world facts. To further probe the reasoning boundaries of T2I models, subsequent research has introduced more specialized benchmarks such as R2I-Bench~\citep{chen2025r2i} and T2I-ReasonBench~\citep{sun2025t2i}.

\section{Details of Rejection Sampling}
We use BAGEL's own understanding module to filter the generated images. The specific instruction prompt is in \autoref{tab:reject_prompt}.

\begin{table*}[!ht]\centering
\begin{minipage}{0.95\textwidth}
    \centering
    \small
    \begin{tcolorbox}[colback=white, colframe=black, boxrule=0.5pt, arc=0pt, left=4mm, right=4mm, top=2mm, bottom=2mm, title=\textbf{Prompt: Strict Image Quality Filter}]
        \textbf{System Instruction:}
        You are an expert image quality assessor. Your task is to evaluate an image based on a specific question. You must be extremely strict and analytical in your evaluation.

        \textbf{Question:} \texttt{\{question\}}

        \textbf{Please follow these steps exactly:}
        \begin{enumerate}[leftmargin=*]
            \item \textbf{Analyze the Question}:
            \begin{itemize}
                \item First, identify the \textbf{type of object(s)} mentioned in the question (e.g., desks, chairs, cats, etc.).
                \item Next, determine the \textbf{correct number} of objects required by the question. If the question contains a mathematical expression (e.g., `8 / 4', `5 - 3'), you MUST perform the calculation first to get the correct number.
            \end{itemize}

            \item \textbf{Examine the Image and Count Objects}:
            \begin{itemize}
                \item Carefully examine the image and count the number of objects of the identified type.
                \item Describe the objects you see and state the actual count.
            \end{itemize}

            \item \textbf{Perform a Strict Criteria Check}:
            \begin{itemize}
                \item Based on your analysis and count, check if the image meets \textbf{ALL} of the following criteria. \textbf{Be very strict.}
                \item \textbf{Correct Number}: The actual count of objects in the image \textbf{MUST match} the correct number you determined in step 1.
                \item \textbf{Correct Type}: All objects found must be of the correct type as specified in the question.
                \item \textbf{No Extra Objects}: The image should not contain any other objects that are not mentioned in the question.
                \item \textbf{Clear Quality}: The image must be clear, recognizable, and free from blurriness or distortion.
                \item \textbf{Complete Objects}: The objects must be complete and uncut, with no parts missing or obscured.
            \end{itemize}

            \item \textbf{Final Judgment}:
            \begin{itemize}
                \item After completing the checks in step 3, provide your final judgment using \textbf{ONLY} one of the two formats below. Do not add any extra text or explanation.
                \item If all criteria are met: \texttt{Final Answer: YES}
                \item If even ONE criterion is NOT met: \texttt{Final Answer: NO}
            \end{itemize}
        \end{enumerate}
        Think step by step and be very strict in your evaluation.
    \end{tcolorbox}
    \vspace{-2mm}
    \caption{\textbf{Prompt for Rejection Sampling.} This prompt is used to rigorously filter generated images by verifying object counts, types, and visual quality against the input prompt.}
    \label{tab:reject_prompt}
\end{minipage}
\end{table*}

\end{document}